\RequirePackage{fix-cm}
\documentclass[twocolumn]{svjour3}          
\usepackage{graphicx}
\usepackage{multicol} 
\usepackage[labelfont=bf]{caption}

\smartqed  
\usepackage{graphicx}
\usepackage{orcidlink}
\usepackage{cite}
\usepackage{amsmath,amssymb,amsfonts}
\usepackage{algorithmic}
\usepackage{graphicx}
\usepackage{textcomp}
\usepackage{multirow}
\usepackage{hyperref}
\usepackage{enumitem}
\usepackage{float}
\usepackage{subcaption}
\usepackage{amsmath}  
\usepackage{lipsum}  
\usepackage{wrapfig}
\usepackage{xurl}
\usepackage{graphicx}
\usepackage{lipsum}  

\usepackage{lipsum}                     
\usepackage{xargs}                      
\usepackage[colorinlistoftodos,prependcaption,textsize=tiny]{todonotes}
\newcommandx{\unsure}[2][1=]{\todo[linecolor=red,backgroundcolor=red!25,bordercolor=red,#1]{#2}}
\newcommandx{\change}[2][1=]{\todo[linecolor=blue,backgroundcolor=blue!25,bordercolor=blue,#1]{#2}}
\newcommandx{\info}[2][1=]{\todo[linecolor=OliveGreen,backgroundcolor=OliveGreen!25,bordercolor=OliveGreen,#1]{#2}}
\newcommandx{\improvement}[2][1=]{\todo[linecolor=Plum,backgroundcolor=Plum!25,bordercolor=Plum,#1]{#2}}
\newcommandx{\thiswillnotshow}[2][1=]{\todo[disable,#1]{#2}}

\usepackage{xcolor}
\def\BibTeX{{\rm B\kern-.05em{\sc i\kern-.025em b}\kern-.08em
    T\kern-.1667em\lower.7ex\hbox{E}\kern-.125emX}}

\newlist{questions}{enumerate}{2}
\setlist[questions,1]{label=\textbf{RQ}\arabic*.,ref=RQ\arabic*}
\setlist[questions,2]{label=(\alph*),ref=\thequestionsi(\alph*)}

\usepackage{ifsym}
\usepackage{marvosym}

\begin{document}

\title{Prescriptive Process Monitoring Under Resource Constraints: A Reinforcement Learning Approach
}



\newcommand\Mark[1]{\textsuperscript#1}
\author{Mahmoud Shoush\Mark{1}\orcidlink{0000-0002-7423-9909}  \and Marlon Dumas\Mark{1}\orcidlink{0000-0002-9247-7476}}

\institute{\Letter\hspace{2mm} Mahmoud Shoush \at
              \makebox[0pt][l]{\hspace{6mm}\href{mailto:mahmoud.shoush@ut.ee}{\texttt{mahmoud.shoush@ut.ee}}}             
           \and
           \makebox[0pt][l]{\hspace{1mm}}\phantom{Marlon Dumas\hspace{-15mm}} Marlon Dumas \at
              \makebox[0pt][l]{\hspace{6mm}\href{mailto:marlon.dumas@ut.ee}{\texttt{marlon.dumas@ut.ee}}}   \\
           \at
           \begin{tabular}{@{}l l l@{}}
    \Mark{1} & Institute of Computer Science, University of Tartu, \\
             & Tartu, Estonia
  \end{tabular}         
}

\date{Received: \today  / Accepted: a much later date}

\maketitle

\begin{abstract}
Prescriptive process monitoring methods seek to optimize the performance of business processes by triggering interventions at runtime, thereby increasing the probability of positive case outcomes. These interventions are triggered according to an intervention policy. Reinforcement learning has been put forward as an approach to learning intervention policies through trial-and-error. Existing approaches in this space assume that the number of resources available to perform interventions in a process is unlimited – an unrealistic assumption in practice. This paper argues that, in the presence of resource constraints, a key dilemma in the field of prescriptive process monitoring is
to trigger interventions based not only on predictions of their necessity, timeliness, or effect but also on the uncertainty
of these predictions and the level of resource utilization. Indeed, committing scarce resources to an intervention when the necessity or effects of this intervention are highly uncertain may intuitively lead to suboptimal 
policies. Accordingly, the paper proposes a reinforcement learning approach for prescriptive process monitoring that leverages conformal prediction techniques to consider the uncertainty of the predictions upon which an intervention decision is based. An evaluation using real-life datasets demonstrates that explicitly modeling uncertainty using conformal predictions helps reinforcement learning agents converge quickly towards policies with higher net intervention gain.

\end{abstract}


\section{Introduction}

Prescriptive process monitoring (PrPM) is a family of methods to optimize the execution of business processes by recommending or triggering designated actions during the execution of a case (herein called \emph{interventions}) that are expected to increase the probability of the case ending in a positive outcome~\cite{DASHTBOZORGI2023102198,shoush2021prescriptive}. For example, in a loan origination process, a PrPM method may recommend to a loan officer (the \emph{resource}) to prepare and send a second loan offer to an applicant (the \emph{intervention}) to increase the probability that the applicant will accept at least one of the offers (the \emph{positive outcome}). The goal of such interventions is to maximize a given \emph{gain function} that considers the benefits of producing additional positive case outcomes thanks to the interventions, minus the cost of the interventions themselves.

Most existing PrPM approaches make the assumption that the number of resources available to perform interventions in a process is unlimited~\cite{ detiming,metzger2020triggering,bozorgicaise2023}. In practice, though, an intervention requires a time commitment from scarce resources, such as, for example, a time commitment from a loan officer to prepare a second loan offer in a loan origination process. 
In this setting, every time a PrPM approach triggers an intervention in a given state of a case, it locks in a resource that is no longer available for a while, known as \textit{intervention duration}, to perform other potentially higher-gain interventions.
Thus, in the presence of limited resources, a PrPM approach needs to trigger interventions in a way that takes into account not only the expected gain of this intervention decision. Additionally, it should account for the current level of resource utilization, the time window during which this intervention may occur, and the uncertainty of the factors upon which the intervention decision is taken, e.g., significance and urgency.

An \emph{intervention policy} drives the decision to trigger interventions in a PrPM approach.
Several studies have shown that Reinforcement Learning (RL) is an effective approach to learning intervention policies for PrPM~\cite{DBLP:conf/caise/PalmMP20,metzger2020triggering,bozorgicaise2023}. However, these approaches do not consider resource limitations.
In this setting, we address the following research question: \textit{How can online reinforcement learning be effectively incorporated into a PrPM approach to optimize resource allocation, ensure timely interventions, quickly converge, and maximize a total gain function within resource limitations and intervention uncertainties?} 



To address this question, we introduce an online RL-based approach for PrPM that uses predictive and causal models to estimate the necessity of an intervention and the expected impact of the intervention, coupled with a survival model for estimating the urgency of intervention and conformal prediction techniques to factor in uncertainty. Additionally, the proposed approach includes a capacity-monitoring component designed to keep track of resource utilization and demand intensity to incorporate these parameters into the intervention policy.

We report on an empirical evaluation to assess the effectiveness of different variants of the proposed approach, particularly in terms of convergence and performance metrics across various resource utilization levels. This evaluation is compared to baseline methods that are RL-based but lack consideration for resource constraints and inherent uncertainty parameters within the policy. The objective is to explain how the RL agent achieves quick convergence towards policies that maximize total gain post-convergence.

The paper is structured as follows: Sect.~\ref{sec:relatedWork} covers related work, Sect.~\ref{sec:approach} outlines our proposal, Sect.~\ref{sec:evaluation} presents and analyzes the empirical evaluation, and Sect.~\ref{sec:conclusion} concludes the study while suggesting future work directions.



\section{BACKGROUND AND RELATED WORK} \label{sec:relatedWork}

The field of PrPM and RL has attracted growing attention from the research community in recent years, with a multitude of methods being proposed and evaluated. These methods can be broadly classified into three distinct categories, as outlined by~\cite{kubrak2022prescriptive}.  

The first category is concerned with optimizing a Key Performance Indicator (KPI) through the guidance or recommendation of the best next activities to perform~\cite{BranchiLearninigToAct,weinzierl2020prescriptive,LeoniDR20,GrogerSM14}, thereby focusing on refining control flow. The second category seeks to optimize resource allocation strategies, e.g., which resource should perform the next activity~\cite{park2019prediction,sindhgatta2016context,abdulhameed2018resource}. Lastly, the third category seeks to minimize the number of cases resulting in a negative outcome. This category, notably, considers control flow dynamics and resource allocation considerations. 

This paper focuses on the third category (minimizing the occurrence of negative outcomes). Within this category, prior studies~\cite{bozorgicaise2023,Metzger2023,Donadello2022,shoush2021prescriptive,metzger2020triggering,fahrenkrog2021fire} can be subdivided into rule-based and RL-based approaches.




Rule-based approaches, such as the one proposed by Fahrenkrog et al.~\cite{fahrenkrog2021fire}, apply predictive models trained on historical data to estimate the probability of negative case outcomes. In our prior work~\cite{shoush2021prescriptive, shoush2022intervene}, we expanded this latter approach by incorporating a causal model to determine intervention effects ($CATE$: conditional average treatment effect) and a resource allocator for monitoring resource usage (i.e.,\ the number of resources that are busy executing interventions). In~\cite{shoush2023conformal}, we further expanded this approach with a conformal prediction technique. Conformal prediction techniques produce a \emph{prediction set} (i.e.,\ possible outcome sets) instead of always predicting one single outcome per case. Hence, the actual outcome is guaranteed to be included in the prediction set with a given confidence level. In~\cite{Donadello2022}, the Authors introduce a PrPM approach that prescribes to users temporal relations among interventions that have to be held or discarded in order to achieve a positive outcome. 

The above approaches take predictions generated by models trained on historical data and define policies by setting thresholds (or defining rules) over these predictions. For example, triggering an intervention whenever the predicted probability of a negative case outcome is above 80\% or whenever the conformal prediction set includes the negative outcome with a confidence of 90\%. These methods presuppose that by setting a threshold at some level and triggering interventions whenever the threshold is exceeded, the total gain of the interventions will be maximized in all situations. In practice, an intervention policy may need to consider further factors to maximize total gain. Factors such as the urgency of intervention and resource utilization levels could collectively impact the overall effectiveness and efficiency of the intervention policy.



In contrast, RL-based approaches, exemplified by~\cite{Metzger2023,bozorgicaise2023,weytjens2021learning,metzger2020triggering}, utilize RL techniques to learn intervention policies. In~\cite{metzger2020triggering,Metzger2023}, an RL agent learns an intervention policy through a number of parameters, such as prediction scores and reliability estimates, while the trade-off between the earliness and accuracy of intervention is discussed. On the other hand, the works done by~\cite{bozorgicaise2023,detiming} utilize causal effect estimates to learn the policy. In these approaches, the agent learns an intervention policy via trial-and-error, without predefined triggering thresholds or rules. Nevertheless, these approaches are contingent on the assumption of limitless resource availability and the availability of certain causal effects estimates.
In practice, this assumption does not hold. This paper addresses this limitation by proposing a RL approach for PrPM that goes beyond intervention need, timeliness, and effects and considers the uncertainty of predictions and resource utilization level.

\section{Approach} \label{sec:approach}

The proposed approach focuses on the sequential decision-making process within contexts characterized by limited resources and uncertainty. In this context, the decision maker, referred to as the \emph{agent} from an RL perspective, adheres to an \emph{intervention policy} that guides the decision of when and whether to execute an intervention while considering the corresponding rewards associated with the decision’s effectiveness. In the following, we will explain our proposal's conceptual foundation, followed by a detailed description of its components. 

\subsection{\textbf{Factors Influencing Intervention Decisions}}

\begin{figure}[H]
  \centering
  \includegraphics[width=0.98\linewidth]{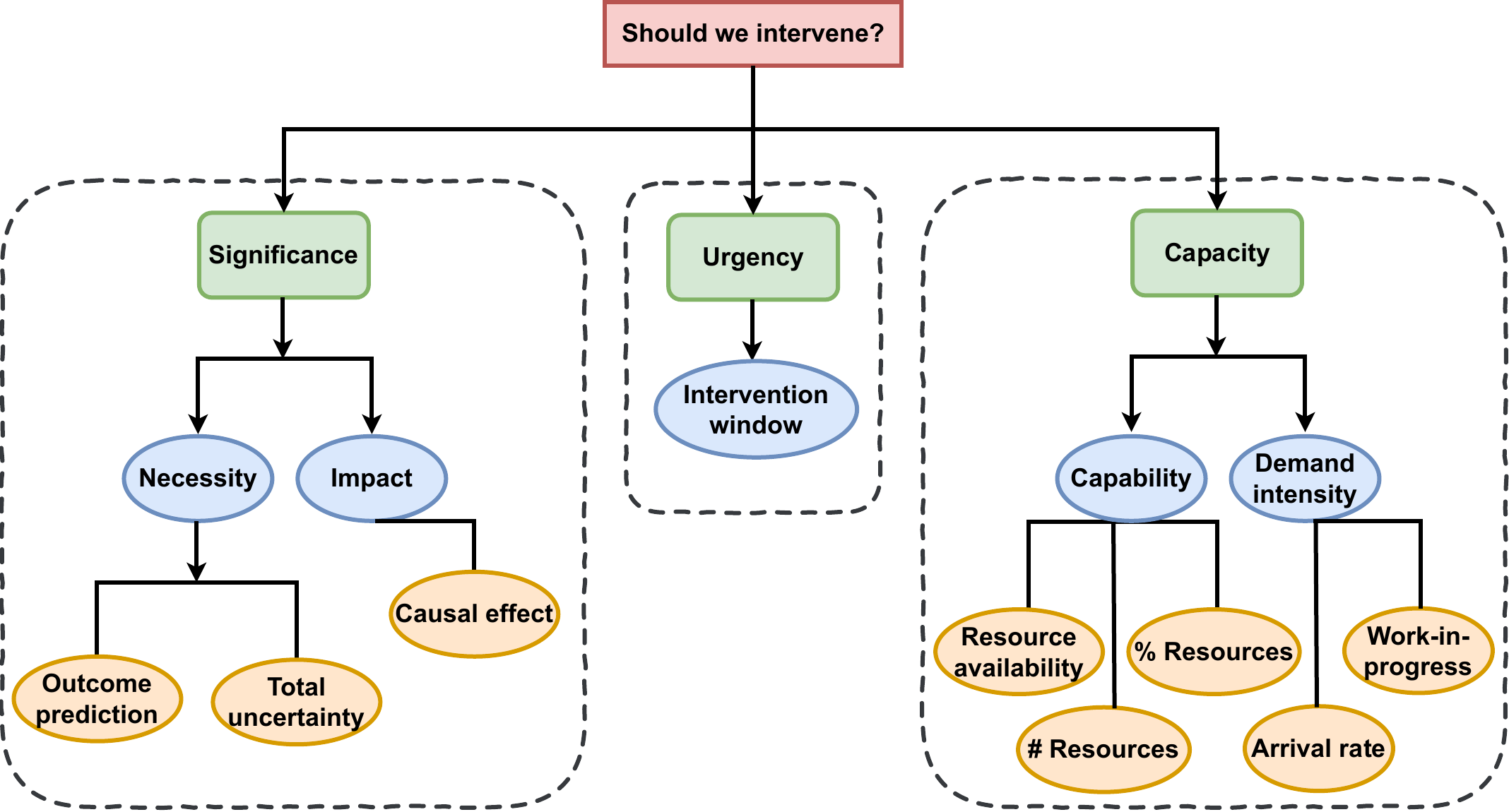}
  \caption{Factors influencing intervention decision-making.}
  \label{fig:dims}
\end{figure}

The conceptual basis of the proposed methodology is designed upon a set of factors. These factors could influence the agent's decision to trigger an intervention and allocate resources to a specific case within a particular state. As illustrated in Fig.\ref{fig:dims}, these factors are categorized into three top-level dimensions: \textit{Significance, Urgency, and Capacity}.

\paragraph{\textbf{Significance}}: The dimension of significance addresses the essential question of whether an intervention has both \textit{necessity} and \textit{impact} for a particular case. As expected, interventions should be triggered when they are supposed to be necessary, signifying that cases are prone to result in a negative outcome. Moreover, these interventions should also hold the potential for significant impact, capable of preventing negative outcomes and thereby delivering value.

The \textit{necessity} of an intervention is determined upon reaching a specific confidence level in predicting a negative outcome for a case. This is justified by the need to ensure that interventions are triggered only when there is a high confidence level in their necessity. To quantify this necessity, we employ a dual-factor approach. The first factor assesses the likely case outcomes (\textit{outcome prediction}), particularly focusing on the probability of a negative outcome. This is essential because we want to identify cases likely to end in a negative outcome, justifying intervention. The second factor accounts for the \textit{total uncertainty} inherent in these predictions. This is crucial as it recognizes that predictions are not always perfect, and varying degrees of uncertainty may be associated with them. We hypothesize that including the total uncertainty helps the RL agent give some importance to some cases compared to others, regardless of their outcome prediction scores. 
A predictive model facilitates this assessment by providing a reliable estimate of the probability of a case likely to end with a negative outcome (outcome prediction) along with the total uncertainty linked to the outcome prediction.

The \textit{impact} of an intervention revolves around the question of how substantially the probability of a positive case outcome is boosted upon triggering the intervention. This is an essential consideration as it directly relates to the effectiveness of the intervention. To measure this impact, we use a causal model to estimate the intervention’s \textit{causal effect} (or treatment effect.) This evaluation is pivotal in determining whether triggering the intervention holds the potential to significantly transform a case with a negative outcome into a positive one. Consequently, providing the RL agent with information regarding the expected impact of triggering the intervention can serve as a guiding principle for the agent to acquire a more optimal intervention policy. For instance, in a customer churn scenario, the causal model determines to what extent a personalized discount offer can prevent a valuable customer from terminating a subscription service.

\paragraph{\textbf{Urgency}} The urgency dimension is primarily concerned with determining the optimal timing for intervention. It revolves around the decision of whether to intervene now, delay intervention to a later time, or abstain from intervening altogether. This dimension encapsulates the duration remaining until the intervention becomes unfeasible, which we term the \textit{intervention window}. Although the RL agent could potentially infer information about the time left to intervene indirectly, we hypothesize that including intervention window information will speed the convergence of the policy, leading to fewer trial-and-error steps and enhancing the agent’s decision-making efficiency, especially when resources are limited.



In situations characterized by resource constraints, consider a scenario where only one resource is allocated for the execution of the intervention, reflecting a high level of resource utilization. For instance, only one loan officer is available to prepare and send the applicant a second loan offer. Simultaneously, this scenario is further complicated by the presence of three cases that urgently necessitate immediate intervention due to their time-sensitive nature. From the significance perspective, it is evident that cases one and two exhibit a higher likelihood of negative outcomes and marginally more significant greater causal impact than case three. However, according to the intervention window information, case three is expected to end with a negative outcome much sooner than cases one and two, hence, less time to intervene. This temporal urgency needs immediate attention. Thus, an optimal policy may opt to postpone interventions for cases one and two while allocating the available resources to address case three, guided by the insights from the intervention window. Thus, an ML model capable of predicting the forthcoming negative outcome in case three while suggesting deferral of interventions for cases one and two becomes essential. Survival models~\cite{DBLP:conf/icpm/Baskharon0F20}, which specialize in estimating the time duration until a specific event, as a negative outcome occurs, offer a fitting approach for estimating the intervention window. This strategy could potentially optimize resource utilization under constraints and maximize the number of cases to end with a positive outcome. 

\paragraph{\textbf{Capacity}} The dimension of capacity plays a pivotal role in evaluating the feasibility of interventions. Its essence lies in the fact that triggering an intervention for a given case in a particular state may be necessary, effective, and urgent. However, it may become unfeasible if resources are insufficient or the demand for intervention exceeds the available capacity. In such instances, the trigger of the intervention becomes impractical. Thus, we categorize this dimension into \textit{capability} and \textit{demand intensity}. This categorization helps in a more comprehensive understanding of the capacity dimension.


\emph{Capability} This sub-dimension relates to the ability and feasibility of executing an intervention, which depends on the availability of sufficient resources. It also reflects the extent to which the available resources are utilized. In situations where these resources are lacking, the feasibility of the intervention becomes restricted. To effectively model this, we consider three factors: a boolean variable that signifies the availability of resources, the actual number of available resources denoted as ($n$), and the proportion of resources available represented by ($\eta$). Incorporating these factors is well-justified as it ensures a comprehensive estimate of the capability of executing interventions, containing both binary availability and quantitative factors. This holistic approach collectively influences the feasibility of triggering interventions, which is essential for informed decision-making in resource-constrained environments.



\emph{Demand intensity}, the other sub-dimension, considers situations where there may be sufficient resources for an intervention, but the decision to intervene could be delayed. For example, if there are three available resources and three cases requiring intervention, utilizing all resources would consume the entire intervention resource pool. This decision would result in having no resources left for potential future cases that might arrive shortly and could potentially achieve higher gains from executing the intervention. To tackle this issue, we propose incorporating two factors into our approach: the arrival rate ($\lambda$) of cases and the work-in-progress ($WIP$). As the number of incoming cases increases, the demand for available resources for intervention execution also rises significantly, resulting in a high workload.



The arrival rate gives an estimation of how frequently cases requiring intervention arrive, acting as a real-time indicator of the intensity of arrival. On the other hand, the $WIP$ factor is vital in estimating how many cases are awaiting in the queue and demand intervention. These factors proxy the demand intensity and are crucial in resource allocation optimization and boosting planning ahead. 

For instance, in a practical scenario, such as an IT helpdesk context, the arrival rate reflects the frequency of incoming tickets and directly influences workload management. Conversely, the WIP factor expects future intervention demands, enabling proactive resource allocation. Including these factors could guide the RL agent during the learning process on optimizing resource allocation. 

\begin{figure}[H]
  \includegraphics[width=\linewidth]{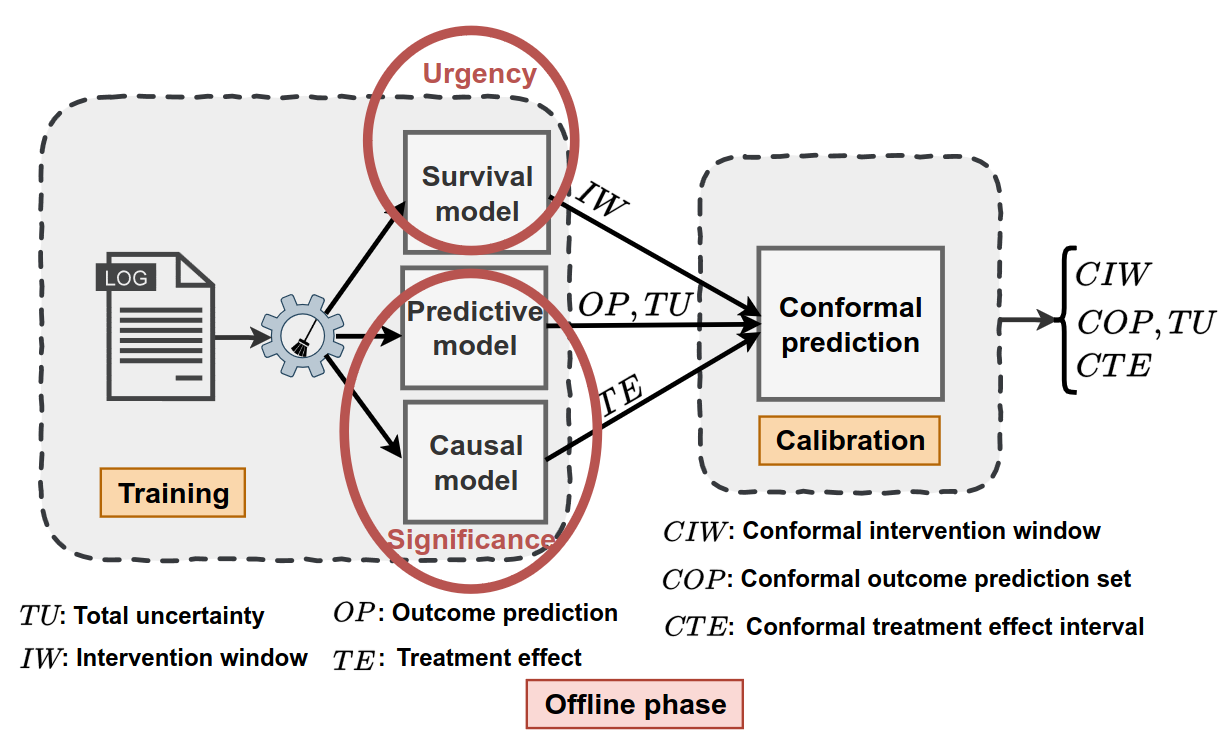}
  \caption{An overview of the offline phase.}
  \label{fig:offline}
\end{figure}

Following the articulation of the dimensions of significance, urgency, and capacity, along with their corresponding factors influencing intervention decisions, we will explain our two-phase approach. This approach consists of an offline phase (depicted in Fig.~\ref{fig:offline}) dedicated to training and calibration. Additionally, an online phase (shown in Fig.~\ref{fig:online}) is specifically designed for real-time learning, during which the RL agent learns the intervention policy through trial-and-error. Both of these phases substantially contribute to our approach’s overall effectiveness. The following subsections will provide a more detailed explanation of the importance and contributions of each phase.

\subsection{\textbf{Offline Phase}}\label{sec:training}

\subsubsection{\textbf{Event Log Preprocessing}}

Data preparation is crucial in predictive and prescriptive tasks. In this context, data preparation includes data cleaning, case prefix extraction, and feature encoding. In the data cleaning step, we follow.~\cite{teinemaa2019outcome} to remove incomplete traces from the event log and events with incorrect timestamps. Then, we extract length prefixes from each case, reflecting real-world scenarios where we need to make intervention decisions on incomplete cases. In other words, each prefix denotes a decision point, or a state for the RL agent, that requires deciding whether to allocate resources and execute an intervention (or not). 


The third step is to map each prefix of a case to a feature vector. This is necessary to train machine learning models.
This feature encoding step is achieved as follows:
\begin{itemize}
\item Each case attribute\footnote{In an event log, a case attribute is an attribute that takes the same value throughout the case, in other words, it is inherent to the case. An event attribute is an attribute that may change value between one event in a case and the following one.} is mapped to a feature (numerical or categorical, depending on the datatype of the case attribute). For example, an attribute such as the ``loan type'' is a categorical case attribute. It is thus mapped to a categorical feature.
\item Each event attribute is mapped into a set of features by using the so-called \textit{aggregate encoding} method as defined in~\cite{teinemaa2019outcome}. In this feature encoding approach, if an event attribute is categorical, we introduce one feature for each value of the attribute. For a given value of an event attribute, the corresponding feature is a positive integer indicating how many times that value has occurred in the encoded case prefix. For example, if one of the attributes is ``Resource'' and one of the possible values of this attribute is ``John Smith'', we introduce a feature corresponding to the number of events in the case prefix where the value of the ``Resource'' attribute is ``John Smith''. On the other hand, a numerical event attribute is encoded as a single feature corresponding to the mean of the values of this attribute observed in the prefix. For example, an event attribute ``Payment due'' is mapped to a feature corresponding to the mean of the values of the ``Payment due'' attribute across the events in the case prefix.
\item Previous work on predictive process monitoring has highlighted the importance of including temporal features~\cite{Taxcaise}. Accordingly, We include features to encode timestamp-related details: 'time to last event,' 'time since case start,' 'time since last event,' 'time since midnight,' 'month,' 'weekday,' and 'hour.'
\item The above attributes are computed for each case in isolation. Research in the field of predictive process monitoring has highlighted the importance of introducing features computed across all active cases (as of the most recent timestamp in the case prefix)~\cite{DBLP:journals/is/SenderovichFM19}. According to Little’s Law~\cite{LittlesLaw}, two primary factors impacting process workload are ($WIP$), i.e.\ the number of active cases, and the arrival rate, i.e.\ the number of cases created per time unit. Accordingly, we also include two features corresponding to these two measures.
\item In addition to the above, we may add other log-specific attributes to capture domain-specific requirements. For example, in a loan origination log introduced later in the paper, we include features to encode the changes in two key attributes: changes in the number of loan offers sent to applicants and changes in the monthly interest offered to the applicant. 
\end{itemize}

\subsubsection{\textbf{Survival Model}}\label{sec:survival}
The survival model~\cite{klein2003survival} predicts the time until a specific event, such as a negative outcome, occurs. This analysis is particularly valuable under resource constraints as it helps prioritize queued cases for timely interventions, thereby preventing costly negative events. The objective is to balance the costs associated with intervening either too early or too late, recognizing that timing is crucial for efficient resource allocation. Consequently, we use the survival model to determine an \textit{intervention window} ($IW$) – the timeframe during which an intervention can be implemented before the critical event happens – which gives vital insights into the urgency of decision-making and case prioritization. We explicitly provide the intervention window to the RL agent to accelerate convergence. More, it reduces the need for additional trial-and-error steps compared to allowing the RL agent to infer this information during runtime.

For example, in supply chain management, a survival model offers practical benefits. It can prioritize orders by estimating their expected arrival time, enabling a focus on urgency. Also, estimating the time remaining until orders arrive helps optimize inventory management, reduces stockout risk, and minimizes customer delivery delays. This approach allows timely interventions like speeding orders or increasing production capacity, preventing negative outcomes such as stockouts or delayed deliveries. Consequently, it helps avoid customer dissatisfaction and potential revenue loss.

\subsubsection{\textbf{Predictive Model}}\label{sec:predicitve}
Our earlier work~\cite{shoush2022intervene} detailed a predictive model for PrPM.  This paper gives a concise overview of the model and highlights how we intend to apply it.

The predictive model estimates two critical measures for each case at various prefixes or decision points. These measures are the probability of a negative outcome, denoted as the \emph{outcome prediction} ($OP$), and the quantification of \emph{total uncertainty} ($TU$). The $OP$ is determined by aggregating scores derived from an ensemble of predictions. At the same time, $TU$ is estimated using the entropy of the average prediction, following the approach introduced by~\cite{malinin2020uncertainty}.

The integration of both $OP$ and $TU$ plays a substantial role in influencing triggering intervention at different decision points, enabling the RL agent's evaluation of intervention necessity. To illustrate, when the predictive model gives a high probability of a negative outcome for a case but with a notable degree of uncertainty, the agent may decide against triggering the intervention and allocating resources. Instead, resources could be directed toward another case where the predicted outcome carries a higher level of certainty. Therefore, considering $TU$ introduces varying degrees of importance among different cases, allowing for more informed intervention decisions and allocation of resources.

\subsubsection{\textbf{Causal Model}}\label{sec:causal}
The primary objective of the causal model is to estimate the potential intervention effect~\cite{Pearl10Causal,DBLP:journals/is/BozorgiTDRP23}. Specifically, its impact on reducing the probability of a negative outcome and, accordingly, increasing the probability of a positive outcome. It estimates whether the intervention's effect could effectively influence the probability of a positive outcome in a manner that increases it. This estimation is accomplished by comparing the probabilities of an event occurring with and without the intervention. For instance, if the probability of a purchase order is 0.9 with the intervention and 0.3 without it, the intervention effect is estimated as 0.6. Therefore, quantifying the intervention effect provides crucial insights into the effectiveness of interventions in mitigating negative outcomes.

Different measures, including individual treatment effects  ($TE$) and conditional average treatment effects ($CATE$), can be used to represent the intervention effect. The $TE$ quantifies the difference in outcomes for a given case at a particular decision point when subjected to the intervention compared to when they are not. It allows us to understand how the intervention affects each prefix uniquely. Conversely, $CATE$ focuses on the average effect of an intervention on a subgroup of cases that share similar characteristics or conditions. It involves partitioning the dataset into groups or cohorts based on certain attributes or conditions and then calculating each group’s average treatment effect. Hence, $CATE$ helps us understand how the intervention’s impact varies across different subpopulations, providing insights into its effectiveness under various circumstances. Both $TE$ and $CATE$ are valuable for evaluating the intervention effect. However, we have chosen to focus on TE rather than $CATE$ due to some limitations associated with the latter, as pointed out by~\cite{lei2021conformal}.

Specifically, $CATE$ does not adequately account for individual variations in response to interventions, which is a crucial factor in effective decision-making. Additionally, accurate estimation of $CATE$ often necessitates access to extensive and sometimes hard-to-obtain covariate data. In contrast, $TE$ is a more suitable choice, primarily because it considers data availability constraints and individual response variability. For example, $TE$ allows for a comprehensive assessment of the impact of marketing campaigns on each customer, capturing variations that $CATE$ might overlook. As a result, we hypothesize that $TE$, with its adaptability to practical data limitations and individualized responses, will be a more effective choice for guiding the RL agent in learning an optimal intervention policy.

\subsubsection{\textbf{Calibration}}\label{sec:calibration}
During the calibration step, we use a technique from the conformal inference principles~\cite{shafer2008tutorial,tibshirani2019conformal,DBLPConfromalReview} to generate estimates that come with a guaranteed confidence level. Notably, such a technique is model-agnostic, meaning it can be applied atop any predictive model to transform its estimates into ones with guaranteed confidence levels. This technique depends on a user-defined significance level ($\alpha$) to produce reliable estimates, ensuring confidence guarantees of $1-\alpha$. Consequently, a lower significance level corresponds to a higher level of confidence. For example, if a user wants a confidence of 90\%, then the significance level is set to 0.1.

This calibration step is applied to the predictive, survival, and causal estimates, specifically $OP$, $IW$, and $TE$, to ensure a predetermined confidence level. Through the principles of conformal inference, these estimates are transformed into valid prediction sets for classification tasks, e.g., $OP$, and intervals for regression tasks, e.g., $IW$ and $TE$. Consequently, unlike single-point estimates with low confidence, the RL agent has reliable estimates to make confident decisions.

For example, applying the conformal technique to enhance the predictive model in a classification task like the outcome prediction creates a prediction set that we term \textit{conformalized outcome prediction} ($COP$). This set may include a single outcome like \{negative\} with a confidence level of $1-\alpha$, thus rejecting the alternative with specified confidence. However, if no outcome meets the desired confidence, the set encompasses both outcomes (\{negative, positive\}). If neither outcome meets the set confidence, the prediction set remains empty. This approach provides more reliable estimations than low-confidence single-class labels or probability estimates.

We decided not to implement the calibration step for the $TU$ measure because $TU$ already quantifies prediction uncertainty. Also, the $TU$ score could give more information than the boolean information from the conformal prediction set. Hence, introducing an additional layer of uncertainty measurement, especially in conjunction with the probability of a negative outcome, would enhance the intervention policy. This enhancement would guide the RL agent toward more precise decisions, thereby improving both convergence and performance aspects.

In contrast, when the conformal technique is applied atop the survival model, which is a regression task, it generates a prediction interval referred to as a conformalized \textit{intervention window} ($CIW$). This interval provides a range for the time remaining until a negative outcome occurs. This is crucial because relying solely on a single-time estimate can introduce inaccuracies and affect the timing of interventions. For instance, if the model predicts a negative event in four days but occurs in three, it could result in a delayed intervention, potentially leading to a case loss. In contrast, the $CIW$ offers a prediction interval (e.g., $[2.5:3.5]$ days) with a confidence guarantee, enhancing the estimate’s reliability and facilitating accurate intervention timing.

Similarly, when the conformal technique is applied atop the causal model, it generates a prediction interval referred to as a \textit{conformalized treatment effect} ($CTE$), thus ensuring robust treatment effect estimations. This methodology allows for measuring uncertainty related to the estimated causal effects and guarantees confidence. To explain its usage, consider a scenario where a business is analyzing the impact of a new policy on customer satisfaction. By integrating conformal inference within the causal model, prediction intervals for individual policy impacts can be determined, providing a spectrum of potential effects for each customer. This information helps make well-informed decisions about the policy’s efficacy for different customers while acknowledging the inherent uncertainty in these estimates.

\subsection{\textbf{Online Phase}}\label{sec:online}

\begin{figure}[H]
  \includegraphics[width=\linewidth]{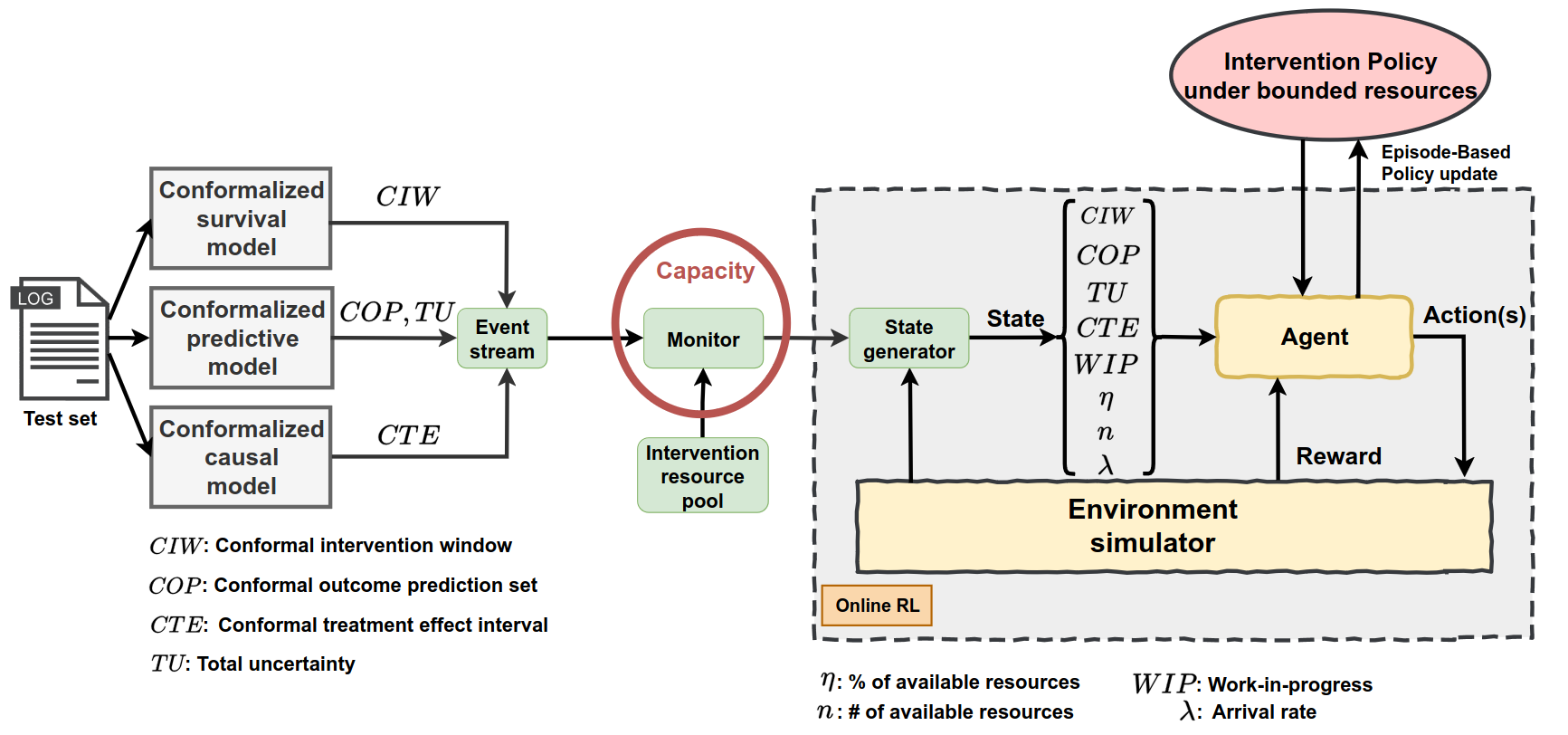}
  \caption{An overview of the online phase.}
  \label{fig:online}
\end{figure}


During this phase, illustrated in Fig.~\ref{fig:online}, we use an online RL algorithm to learn an intervention policy through trial-and-error and operationalize the proposed PrPM approach. To learn this intervention policy, the RL agent interacts with the environment, where resources are limited, and makes real-time decisions at each state within the environment. According to the decisions made by the agent, it receives either rewards or penalties, depending on the quality of its decisions. Over time, it incrementally improves its performance by learning from these experiences and iteratively adjusting its policy, ultimately converging towards the optimal intervention policy. The agent aims to achieve rapid convergence by taking actions that result in higher rewards and fewer penalties, thus maximizing its aggregate performance, represented by the total gain. Through these continuous learning and adaptation processes, the agent converges towards an intervention policy optimized for maximizing the total gain. 

To transform the scenario above into a PrPM task under uncertainty and resource constraints, our first step involves utilizing the models trained during the offline phase. We utilize these models to produce estimates that capture both the significance ($COP$, $TU$, and $CTE$) and urgency ($CIW$) dimensions. These estimates reflect the uncertainty and confidence levels related to the model’s estimates.
Then, we incorporate a \textit{monitoring} component designed to continuously observe event flow, considering arrival patterns and the availability of intervention resources, as defined by domain-specific knowledge. 

This component actively monitors the number ($n$) and percentage ($\eta$) of available resources, the workload represented by cases in progress ($WIP$), and the intervention demand represented by the rate of case arrivals ($\lambda$). 
Additionally, the monitoring component keeps track of the available resources within the intervention resource pool. Hence, when the RL agent triggers the intervention for a specific case, and a resource is available within the intervention resource pool, that resource is allocated to the selected case. Then, the allocated resources are temporarily reserved for a pre-established treatment duration ($T_{dur}$) and released once the allocated time has elapsed. 
This monitoring process significantly contributes to modeling the capacity dimension effectively, ensuring optimal resource planning.

In the following, we will explain the learning problem, including constructing the environment with which the agent interacts. Also, we will provide a detailed explanation of the reward and gain functions.

\subsubsection{\textbf{Learning Problem}}

The dimensions of significance, urgency, and capacity are incorporated into a state generator, resulting in the creation of a state space. This state space helps construct a simulated environment with uncertainty and limited resources. Within this environment, the agent accurately simulates the process execution. Hence, the agent observes a state from the constructed state space at each decision point, effectively simulating the environment. It then chooses an intervention from the available action space, and based on this decision, it is either rewarded or penalized. This process continues as the agent proceeds to the next decision point, observing a new state, and repeating these steps iteratively until the simulation ends.

This iterative process is essential in the agent’s learning process, enabling it to explore and exploit various interventions until it converges towards the optimal intervention policy, which maximizes the total gain. The agent reaches convergence when its decisions align with the criteria of triggering interventions when necessary, being effective, urgent, and when resources are available. This convergence is denoted by continuous positive rewards and gains resulting from these decisions.

The state space contains all relevant information about ongoing cases. Hence, the state is represented by a tuple of factors, including $CIW$, $COP$, $TU$, $CTE$, $WIP$, $\eta$, $n$, and $\lambda$. These factors are included in the state representation to reflect the significance, urgency, and capacity dimensions. Accordingly, this state representation guides the agent in making informed decisions regarding whether and when to trigger interventions and planning ahead.

On the other hand, the action space consists of two potential values: 0, which signifies the decision not to trigger the intervention, and 1, which indicates the choice to trigger it. The intervention refers to any action that positively influences cases likely to end with a negative outcome. We assume that the intervention is predefined based on domain knowledge. For instance, the intervention could be calling a customer to offer a discount or reducing the monthly interest rate on a loan application to increase its chances of approval.

\subsubsection{\textbf{Reward function}}
Deciding whether to trigger the intervention or not can result in rewards when the decision is correct, but it can also lead to penalties when the decision is wrong. Therefore, designing a reward (or penalty) function is crucial in the agent’s learning process. It guides the agent to navigate the complex environment, including both the action and state spaces. This guidance helps the agent identify and take actions that contribute most significantly to optimizing cumulative rewards and total gain.

Therefore, the reward function is carefully designed to incentivize the agent to trigger an intervention when it aligns with significance, urgency, and the presence of available capacity dimensions. While this strategy could result in a policy that maximizes the total gain, it also introduces inherent costs to be considered in the decision-making process. These costs are typically associated with the expense of triggering the intervention, such as the cost of a loan officer’s time. 

Hence, we include several factors in our reward function, such as intervention cost ($C_{in}$), gain from resource allocation ($gain_{res}$), positive outcome gain ($gain_{out}$), and treatment effect (TE). The $gain_{res}$ variable guides the agent to allocate resources when they are available and needed for a case while discouraging resource allocation when the agent triggers an intervention in a situation where resources are unavailable. Similarly, the $gain_{out}$ is used to represent the benefits when a case ends with a positive outcome, scaled by the intervention effect (TE), which directly influences it. As the treatment effect increases, the $gain_{out}$ also rises; conversely, the $gain_{out}$ decreases as the treatment effect decreases. Table~\ref{table:tab1} provides a comprehensive illustration of the reward function under different conditions. 

\begin{table*}[htbp]
\caption{The proposed reward function}
\resizebox{\textwidth}{!}{%
\begin{tabular}{|ll|l|l|ll}
\cline{1-4}
\multicolumn{2}{|l|}{}                                           & \multicolumn{2}{l|}{Agent triggers the intervention}       &  &  \\ \cline{1-4}
\multicolumn{1}{|l|}{Resource available}   & Intervention effect & \multicolumn{1}{l|}{Yes}               & No                &  &  \\ \cline{1-4}
\multicolumn{1}{|l|}{\multirow{3}{*}{Yes}} & Positive            & \multicolumn{1}{l|}{($gain_{out} * TE) - C_{in} + gain_{res}$} & $ - (gain_{out} + gain_{res})$ &  &  \\ \cline{2-4}
\multicolumn{1}{|l|}{}                     & Negative            & \multicolumn{1}{l|}{$ - (gain_{out} + gain_{res} + C_{in})$} & $gain_{out} + gain_{res}$ &  &  \\ \cline{2-4}
\multicolumn{1}{|l|}{}                     & \multirow{2}{*}{Zero} & \multicolumn{1}{l|} {positive outcome: $-(C_{in}+gain_{res})$}  & {positive outcome: $gain_{out}+gain_{res}$} &     \\ \cline{3-4}
\multicolumn{1}{|l|}{}       & \multicolumn{1}{l|}{} & {negative outcome: $-(gain_{out}+C_{in}+gain_{res})$} &  {negative outcome: $0$}  &  \\ \cline{1-4}
\multicolumn{1}{|l|}{\multirow{3}{*}{No}}  & Positive            & \multicolumn{1}{l|}{$- (gain_{res})$} & $gain_{res}$ &  &  \\ \cline{2-4}
\multicolumn{1}{|l|}{}                     & Negative            & \multicolumn{1}{l|}{$- (gain_{res})$} & $gain_{res}$ &  &  \\ \cline{2-4}
\multicolumn{1}{|l|}{}                     & Zero                & \multicolumn{1}{l|}{$- (gain_{res})$} & $gain_{res}$ &  &  \\ \cline{1-4}

\end{tabular}%
}

\label{table:tab1}
\end{table*}

When the agent intervenes, given resource availability, the intervention's effectiveness determines three potential scenarios. For a positive effect, the agent earns a reward, $r = (gain_{out} * TE) - C_{in} + gain_{res}$, for efficiently allocating resources when it is needed and effective while incurring the cost of intervention. Conversely, a negative effect incurs a penalty, $r = - (gain_{out} + gain_{res} + C_{in})$, representing resource misallocation, failure to avoid the negative outcome, and intervention costs. With zero effect, penalties depend on case outcome: $r = -(C_{in}+gain_{res})$ for a positive outcome and $r = -(gain_{out}+C_{in}+gain_{res})$ for a negative one, both reflecting wasted resources, no gains, and intervention costs.

On the contrary, when the agent refrains from triggering the intervention, three potential scenarios unfold depending on the intervention's effectiveness. A positive effect incurs a penalty, $r = - (gain_{out} + gain_{res})$, representing a missed positive outcome. Conversely, a negative effect secures a reward, $r = gain_{out} + gain_{res}$, for avoiding resource misallocation and intervention costs. In situations with a neutral effect, rewards are $r = gain_{res} + gain_{out}$ for a positive outcome and $r = 0 $ for a negative one, implying effective resource utilization.

The above reward function assumes that for each case, we know the outcome of the case if the agent performs an intervention, as well as the outcome of a case if the agent does not perform an intervention. In this way, whichever decision the agent takes (to intervene or not to intervene) we are able to calculate a reward. In real-life event logs, we only know, for each case, one of the two possible outcomes: (1) if an intervention occurred in a case, we know the outcome given that the intervention occurred, but we do not know the outcome should the intervention not be performed; and (2) if an intervention was not performed in a case, we know the outcome given that the intervention occurred, but not the outcome should the intervention had been performed.
To address this lack of information, and inspired by~\cite{DBLP:journals/is/BozorgiTDRP23}, we use an estimator of alternative outcomes, namely RealCause~\cite{DBLP:journals/corr/abs-2011-15007}. Given a dataset where only one possible outcome is known per sample (i.e.\ for each case we either know the outcome given that an intervention occurred or the outcome given that the intervention did not occur), RealCause estimates what the alternative outcome assuming that case outcomes follow a Bernoulli distribution.

In situations where resources are unavailable, we assume interventions cannot be triggered due to resource limitations. Hence, the negative outcome cannot be prevented. We assign rewards or penalties based on resource allocation to guide the agent in these situations. If the agent triggers an intervention, regardless of its effectiveness, it suffers a penalty $r = -(gain_{res})$, marking the wasted effort due to resource constraints. In contrast, if the agent refrains from triggering the intervention, it receives a reward equal to $r = gain_{res}$, rewarding the agent’s choice to refrain from intervention where resources  are unavailable. 

\begin{equation}
        gain = gain_{out} \cdot outcome - frequency \cdot C_{{in}} \label{eq:gain}
\end{equation}



Finally, we estimate the agent’s episode gain, which is updated after each episode. As events progress chronologically, an episode begins when the first event of a given case is observed and finishes when the last event of the same case is observed. We then update the policy at the end of each episode (or case), effectively avoiding any potential data leakage. The gain per episode is then defined as per Eq.~\ref{eq:gain}. 

The gain function considers $gain_{out}$, representing the benefit of achieving a positive outcome at the end of a specific case. This benefit is adjusted based on the observed outcome at the end of that case. If the outcome is positive, the gain reaches its maximum potential; otherwise, there is no gain for the positive outcome. Additionally, this gain is reduced by how many times (or \textit{frequency}) the agent triggered the intervention for that case multiplied by the intervention cost ($C_{in}$). 

\begin{equation}
        Total_{gain} = \sum_{i=1}^c gain_i
        \label{eq:tgain}
\end{equation}

As the simulation reaches its end, the $total_{gain}$ of the agent’s policy is estimated, as illustrated in Eq.~\ref{eq:tgain}, where $c$ denotes the total number of cases encountered throughout the simulation. This estimation is determined by the sum of gains accrued from each case or episode, thus providing a comprehensive measure of the agent’s overall performance. Decoupling the reward function from the total gain is undertaken for several convincing reasons. First, this decision was necessitated by resource constraints, wherein the agent was required to consider available resource allocation effectively. Any misallocation or incorrect utilization of these resources is subject to penalization. Additionally, when the reward function is tied directly to total gain, the agent might hesitate to explore less familiar or riskier decisions. By separating the two, the agent could explore more freely, as the gain calculation remained unaffected by these exploratory actions.

\section{Evaluation} \label{sec:evaluation}
In this section, we demonstrate the effectiveness and applicability of the proposed PrPM approach. In particular, in terms of convergence and performance metrics across various resource utilization levels. This evaluation is compared against RL-based baseline approaches~\cite{bozorgicaise2023,metzger2020triggering} that do not incorporate the factors of resource constraints and the inherent uncertainty parameters within the intervention policy.

The \textit{convergence} point is identified as the moment when the RL agent’s gain becomes positive, signifying its rise above zero, and this state persists unaltered until the end of the simulation. This metric measures the rapidity or the number of cases required for the RL agent to acquire an intervention policy that consistently yields positive gains. In essence, it assesses the agent’s ability to allocate resources to cases when they are available and of significant and critical importance. 

The \textit{performance} is evaluated by considering the total gain achieved upon reaching the convergence point. This choice is driven by the fact that the RL agent requires time to learn the optimal policy for achieving positive gains. Evaluating the approaches during the learning process is impractical. One can view the total gain before convergence as a warm-up phase for the agent, whereas after convergence, the learned policy can be effectively applied in production.



Accordingly, our evaluation aims to explain how the RL agent achieves quick convergence towards policies that maximize total gain post-convergence at different resource utilization levels. Particularly, we address the following research questions: 
    \begin{questions}
        \item How do different variants of the proposed approach perform in terms of both convergence, measured by the number of cases, and performance, measured by total gain post-convergence, across various resource utilization levels?\label{rq:rq1}
        
        
        
        \item How does the performance of a particular variant of our proposal compare to baseline methods regarding convergence and total gain across different resource utilization levels?\label{rq:rq2}
        
        

    \end{questions}






\subsection{\textbf{Datasets}}
\label{subsec:datasets}

\begin{table*}[hbtp]
\caption{Event logs statistics}
	\label{tab:dataset}
 \centering
	\resizebox{0.95\linewidth}{!}{
\begin{tabular}{cccccccc}
\hline
Log           \hspace{2mm}        & \hspace{2mm} \# Cases   \hspace{2mm}      & \hspace{2mm}\begin{tabular}[c]{@{}c@{}}\# Events\end{tabular} \hspace{2mm}& \hspace{2mm}\begin{tabular}[c]{@{}c@{}}Mean\\ length\end{tabular} \hspace{2mm}&  \hspace{2mm}\begin{tabular}[c]{@{}c@{}}Last\\ activity\end{tabular}      \hspace{2mm}       & \hspace{2mm}Outcome  \hspace{2mm} & \hspace{2mm}\begin{tabular}[c]{@{}c@{}}Intervention\\ activity\end{tabular}\hspace{2mm} \\ \hline
\multirow{2}{*}{BPIC2012} & \multirow{2}{*}{$4,688$} & \multirow{2}{*}{$115,125$}                                  & \multicolumn{1}{c|}{\multirow{2}{*}{$24$}}            & A\_pending                                                          & positive   & -                                                               \\ \cline{5-7} 
                          &                         &                                                      & \multicolumn{1}{c|}{}                                & \begin{tabular}[c]{@{}c@{}}A\_Canceled \\ A\_Declnied\end{tabular} & negative & Creat\_Offer                                                    \\ \hline
\multirow{2}{*}{BPIC2017} & \multirow{2}{*}{$31,411$} & \multirow{2}{*}{$1,198,319$}                                  & \multicolumn{1}{c|}{\multirow{2}{*}{$38$}}            & A\_Approved                                                         & positive   & -                                                               \\ \cline{5-7} 
                          &                         &                                                      & \multicolumn{1}{c|}{}                                & \begin{tabular}[c]{@{}c@{}}A\_Canceled \\ A\_Declnied\end{tabular}  & negative & Creat\_Offer                                                    \\ \hline
\multirow{2}{*}{TrafficFines} & \multirow{2}{*}{$129,615$} & \multirow{2}{*}{$519,585$}                                  & \multicolumn{1}{c|}{\multirow{2}{*}{$5$}}            & Payment                                                         & positive   & -                                                               \\ \cline{5-7} 
                          &                         &                                                      & \multicolumn{1}{c|}{}                                & \begin{tabular}[c]{@{}c@{}}Send for\\ Credit Collection\end{tabular}  & negative & Add penalty                                                    \\ \hline
\end{tabular}
}
\label{tab:dataset}
\end{table*}

In our experiments, we utilized three publicly available real-life event logs\footnote{\url{https://doi.org/10.5281/zenodo.8352841}}. Among these logs, \textit{BPIC2017}\footnote{\url{https://doi.org/10.4121/uuid:5f3067df-f10b-45da-b98b-86ae4c7a310b}} and \textit{BPIC2012}\footnote{\url{https://data.4tu.nl/articles/dataset/BPI_Challenge_2012/12689204/1}} are derived from the banking sector and refer specifically to a loan origination process. The third log, \textit{TrafficFines}\footnote{\url{https://data.4tu.nl/articles/dataset/Road_Traffic_Fine_Management_Process/12683249}}, corresponds to a road traffic fine management process. As explained in Table~\ref{tab:dataset}, these logs offer a diverse range of case and event numbers, enhancing our evaluation's robustness. For example, the RL agent encounters fewer cases in the case of \textit{BPIC2012}, which features a relatively smaller number of cases and events than the other two logs. Consequently, the total gain post-convergence could be smaller than for the other two logs. 
The \textit{trafficFines} log has a significantly shorter minimum mean case length than the others, which can influence the behavior of the RL agent, potentially affecting its learning dynamics and strategy adaptation. 

The outcome of a case determined by a condition (boolean function) evaluated on a completed case, as detailed in Table~\ref{tab:dataset}. For example, in the TrafficFines log, we categorize a case as having a negative outcome when the fine remains unpaid, resulting in its referral to a credit collection agency.
To determine if an intervention occurs or not in a case, we designate one of the activities in the log as the \emph{intervention activity}. For example, in the TrafficFines log, we designate the ``Add penalty'' activity as the intervention activity. In other words, we assume that placing a penalty to an unpaid fine, increases the probability that the fine is paid. 

Within the loan origination process, the negative outcome is defined as the application being either rejected by the applicant or canceled by the bank. The intervention activity is to make an additional loan offer to the applicant to enhance the probability of the client accepting a loan offer.

\subsection{\textbf{Experimental Setup}}

In our experimental setup, we used Python version 3.8. First, we split each log into three subsets: training (50\%), calibration (25\%), and testing (25\%). We use the training and calibration sets during the offline phase. The training set is used to train predictive, causal, and survival models, while the calibration set is used to obtain estimates with a guaranteed level of confidence. In contrast, the testing set is used during the online phase, simulating the environment for the agent to learn the intervention policy.


The predictive model is trained through an ensemble approach~\cite{malinin2020uncertainty} employing the Gradient Boosting Decision Tree (GBDT) algorithm, specifically \textit{CatBoost}~\cite{prokhorenkova2018catboost}. This model is designed to estimate both the probability of a negative outcome and the total uncertainty. In parallel, the causal model is trained using the \textit{Orthogonal Random Forest} (ORF) algorithm from the \emph{EconML}\footnote{\url{https://github.com/microsoft/EconML}} library to estimate the treatment effect. These methods were chosen due to their demonstrated accuracy and effectiveness in previous studies~\cite{teinemaa2019outcome,shoush2022intervene} .


Furthermore, the survival model is trained following the \textit{Cox proportional hazards}~\cite{fox2002cox} method, a widely adopted statistical model in survival analysis. This method investigates the association between covariates (independent variables) and the hazard rate, which denotes the risk of an event occurring over time. This method is suitable for analyzing time-to-event data, including scenarios like the time until a negative outcome occurs. To implement this method, we used the \emph{lifelines}\footnote{\url{https://github.com/CamDavidsonPilon/lifelines/tree/master}} Python library. This library has been designed explicitly for survival analysis, offering a rich toolkit that enhances the precision and reliability of our survival estimates.

\begin{table*}[htbp]
\caption{The Resource Utilization levels table provides a summary of the parameters recorded after the simulation, their association with resource utilization ($\rho$) levels, and the number of available resources ($n$) employed during the simulation.}
\resizebox{\linewidth}{!}{
\begin{tabular}{ccccccccll}
\cline{1-8}
\multicolumn{4}{c|}{\textbf{}}                                                                                                                                                                                                                                    & \multicolumn{4}{c}{Resource utilization ($\rho$)}                                                                                                                                                                                                                                                                                                                        &  &  \\ \cline{1-8}
\multicolumn{1}{c|}{\textbf{Event log}} & \begin{tabular}[c]{@{}c@{}}\# Tirggered \\ interventions\end{tabular} & \begin{tabular}[c]{@{}c@{}}$T_{dur}$ \\ (s)\end{tabular} & \multicolumn{1}{c|}{\begin{tabular}[c]{@{}c@{}}Duration of \\ the log (s)\end{tabular}} & \begin{tabular}[c]{@{}c@{}}High\\ $\rho$ $\geq$ 90\%\end{tabular} & \begin{tabular}[c]{@{}c@{}}Moderately High\\ 90\% \textgreater $\rho$ $\geq$ 75\%\end{tabular} & \begin{tabular}[c]{@{}c@{}}Medium\\ 75\% \textgreater $\rho$ $\geq$ 50\%\end{tabular} & \begin{tabular}[c]{@{}c@{}}Low\\ 50\% \textgreater $\rho$ $\geq$ 25\%\end{tabular} &  &  \\ \cline{1-8}
\multicolumn{1}{c|}{BPIC2012}           & 1172                                                                  & 1                                                     & \multicolumn{1}{c|}{365}                                                                & \textbf{1}$\leq$ $n$ $\leq$ 3                                                                     & 3\textless $n$ $\leq$ \textbf{4}                                                                                               & 4\textless $n$ $\leq$ \textbf{6}                                                                                         & 6\textless $n$ $\leq$ \textbf{12}                                                                                      &  &  \\
\multicolumn{1}{c|}{BPIC2017}           & 7852                                                                  & 1                                                     & \multicolumn{1}{c|}{3630}                                                               & \textbf{1}$\leq$ $n$ $\leq$ 2                                                                    & 2\textless$n$ $\leq$ \textbf{3}                                                                                                  & 3\textless $n$ $\leq$ \textbf{4}                                                                                          & 4\textless $n$ $\leq$ \textbf{8}                                                                                      &  &  \\
\multicolumn{1}{c|}{TrafficFines}       & 18352                                                                 & 1                                                     & \multicolumn{1}{c|}{1322}                                                               & \textbf{1}$\leq$ $n$ $\leq$ 17                                                                      & 17\textless $n$ $\leq$ \textbf{18}                                                                                              & 18\textless $n$ $\leq$ \textbf{27}                                                                                         & 27\textless $n$ $\leq$ \textbf{55}                                                                                     &  &  \\ \cline{1-8}
\multicolumn{1}{l}{}                    & \multicolumn{1}{l}{}                                                  & \multicolumn{1}{l}{}                                  & \multicolumn{1}{l}{}                                                                    & \multicolumn{1}{l}{}                                                  & \multicolumn{1}{l}{}                                                                               & \multicolumn{1}{l}{}                                                                      & \multicolumn{1}{l}{}                                                                   &  &  \\
\multicolumn{1}{l}{}                    & \multicolumn{1}{l}{}                                                  & \multicolumn{1}{l}{}                                  & \multicolumn{1}{l}{}                                                                    & \multicolumn{1}{l}{}                                                  & \multicolumn{1}{l}{}                                                                               & \multicolumn{1}{l}{}                                                                      & \multicolumn{1}{l}{}                                                                   &  & 
\end{tabular}
}
\label{tab:resourceUtilization}
\end{table*}

Regarding the conformal model, we have used the Inductive Conformal Prediction (ICP) method~\cite{Vovk15}, as introduced in our previous research work~\cite{shoush2023conformal}. In that work, we utilized the ICP method for a classification task, specifically for outcome prediction. This enabled us to generate conformalized outcome predictions and demonstrate how the prediction set size changes according to tuning different significance levels. In this paper, we adapt this method for a regression task, e.g., the survival model, to obtain a conformalized intervention window. Also, we experiment with a specific significance level value of $\alpha=0.1$, thereby ensuring a high confidence level equivalent to 90\%. Conversely, to derive conformalized treatment effects, we used the \emph{cfcausal}\footnote{\url{https://lihualei71.github.io/cfcausal/reference/conformalIte.html}} library, a developed conformal inference tool in the R programming language. This approach provides reliable estimates for identifying cases likely to end with a negative outcome and determining optimal intervention timings and effects.



In line with prior work~\cite{metzger2020triggering,bozorgicaise2023}, we opt for the \textit{Proximal Policy Optimization} (PPO) algorithm~\cite{PPO} as our chosen online RL algorithm. PPO is widely recognized and employed in RL due to its effectiveness in optimizing policies for continuous control tasks. A notable advantage of PPO lies in its efficient utilization of collected events,  accelerating convergence and enhancing the effective utilization of available data. 


The experiments use a medium-cost-benefit strategy with a $C_{in}/gain_{out}$ ratio of 50\%. Hence, we propose that the value derived from a successful outcome is worth $gain_{out}=\$60$, underlining the considerable benefits of a highly desirable positive outcome. Also, we attach a value of $gain_{res}=\$10$ to the usefulness derived from efficient resource allocation, ensuring the essential yet secondary role of efficiency. These benefits are weighed against an intervention cost of $C_{in}=\$30$, which signifies a meaningful but manageable investment, ensuring a balanced intervention approach that is both beneficial and sustainable. These configurations are subject to variation, contingent upon the specific process and domain knowledge. Hence, we have experimented with other cost-to-benefit ratios. Higher ratios slow the RL agent convergence, while lower ratios yield the opposite effect. Additionally, we conducted experiments with $gain_{res}=\$0$, indicating that no signal was provided to the RL agent for efficient resource allocation. Our findings revealed that RL policy performance improved when $gain_{res}$ exceeded zero.



\subsubsection{\textbf{Resource Utilization}}

\textit{Resource utilization} ($\rho$) quantifies the efficiency of resource allocation and management within the proposed approach. Specifically, resource utilization considers both the changes in demand and the constraints imposed by resource capacity. Wherein demand encapsulates the cumulative resource requisition during the simulation lifetime. This demand is calculated as the summation of the total number of triggered interventions over the simulation period multiplied by the average treatment duration. In parallel, the capacity represents the inherent resource-handling potential of the proposed approach. It is determined by multiplying the number of available resources by the simulation duration. 

Our experimental setup explores four resource utilization levels (see Table~\ref{tab:resourceUtilization}): high, moderately high, medium, and low. As the level of resource utilization increases, the number of available resources decreases.
With these resource utilization levels, we can comprehensively evaluate the robustness and effectiveness of our approach under diverse resource constraints. Importantly, these levels are determined post-simulation for each log. Hence, we record the log duration (or simulation), the number of interventions triggered during that duration, and the average treatment duration, as shown in Table~\ref{tab:resourceUtilization}. This data enables us to establish the resource utilization levels for our analysis.

For instance, the high resource utilization level corresponds to scenarios with limited resources, such as having only one available loan officer. This level provides  insights into the RL agent’s efficiency when the demand for intervention exceeds the available resources. In contrast, at the low resource utilization level, multiple resources could be available for executing interventions at each decision point. However, the RL agent may choose not to utilize them because there is no incremental gain from executing the intervention for a given case.  
The moderately high resource utilization allows us to understand resource optimization in situations where resources are available but not fully exploited. The medium resource utilization level acts as a benchmark, reflecting typical real-world resource management conditions and providing a baseline for performance evaluation.


\subsubsection{\textbf{Variants of the Proposed Approach} }

Our experimentation explored various variants of the proposed approach, aiming to sketch each variant’s unique contribution to the intervention policy. We examined four distinct variants: \textit{all}, \textit{withCATE}, \textit{withoutCIW}, and \textit{withoutTU}.

In the \textit{all} variant, we provide the RL agent with comprehensive information encompassing significance, urgency, and capacity. This particular variant mirrors the approach introduced in this paper.

In contrast, the \textit{withCATE} variant replaces conformalized treatment effect (\textit{CTE}) information with lower and upper \textit{CATE} bounds obtained from research conducted by~\cite{bozorgicaise2023}. We introduce this variant to compare our approach with \textit{TE} against the \textit{CATE} in determining the intervention impact. Therefore, this variation sheds light on how the RL agent adapts its behavior while learning the optimal intervention policy.

Conversely, in the \textit{withoutCIW} variant, we withhold information about the remaining time for intervention and the urgency associated with it from the RL agent. This variant assesses whether explicitly providing \textit{CIW} information to the RL agent yields superior results compared to allowing the agent to deduce intervention timing autonomously.

On the other hand, in the \textit{withoutTU} variant, we intentionally refrain from providing the RL agent with information regarding the total uncertainty quantification linked to the outcome prediction scores. This specific variant is structured to examine whether supplying the RL agent with total uncertainty information enhances its ability to prioritize cases, leading to more effective intervention triggering and resource allocation than other cases.

\subsection{\textbf{Results}}
Here, we show results of evaluating the learned intervention policy across various resource utilization levels, focusing on two fundamental aspects: convergence, denoting the point at which the RL agent consistently makes decisions resulting in positive gains, and performance, which assesses the total gain achieved post-convergence. 

Our methodology comprehensively defines four resource utilization levels. These levels are defined post-simulation, where we systematically vary available resources across different ranges. By conducting this extensive resource range analysis, we identify resource availability thresholds for each utilization level\footnote{Additional findings regarding the RL agent’s performance (total gain) for different resource utilization levels are available in the supplementary material: \url{https://github.com/mshoush/RL-prescriptive-monitoring} and \url{https://doi.org/10.5281/zenodo.8352841}}, as shown in Table~\ref{tab:resourceUtilization}. For example, in the \textit{BPIC2012} log, we allocate $n=1$ for the high resource utilization level, $n=4$ for moderately high, $n=6$ for medium, and $n=12$ for the low level.


In addressing ~\ref{rq:rq1}, many factors can be influential, affecting both the speed at which a convergence point is reached and. Resource utilization is one such factor, as an increase in available resources may extend the time required for the agent to understand the impact of resource saturation due to the higher number of resources available for interventions. Consequently, a longer time might be necessary for the agent to reach convergence. Hence, our interest lies in observing how the agent’s convergence varies across distinct resource utilization levels. 

\begin{table*}[hbtp]
\caption{Convergence speed, measured in terms of the number of cases, for different proposal variants across various resource utilization levels and logs.}
\centering
\resizebox{0.75\linewidth}{!}{
\begin{tabular}{|cc|cccc|}
\hline
\textbf{}                                           & \textbf{}  & \multicolumn{4}{c|}{Resource utilization}                                                                                         \\ \hline
\multicolumn{1}{|c|}{Log}                           & Variant    & \multicolumn{1}{c|}{High}           & \multicolumn{1}{c|}{Moderately High} & \multicolumn{1}{c|}{Medium}         & Low            \\ \hline
\multicolumn{1}{|c|}{\multirow{4}{*}{BPIC2012}}     & all        & \multicolumn{1}{c|}{\textbf{629}}   & \multicolumn{1}{c|}{\textbf{1024}}   & \multicolumn{1}{c|}{1158}           & -              \\ \cline{2-6} 
\multicolumn{1}{|c|}{}                              & withCATE   & \multicolumn{1}{c|}{884}            & \multicolumn{1}{c|}{1132}            & \multicolumn{1}{c|}{\textbf{1136}}  & 1142           \\ \cline{2-6} 
\multicolumn{1}{|c|}{}                              & withoutCIW & \multicolumn{1}{c|}{892}            & \multicolumn{1}{c|}{1159}            & \multicolumn{1}{c|}{-}              & -              \\ \cline{2-6} 
\multicolumn{1}{|c|}{}                              & withoutTU  & \multicolumn{1}{c|}{772}            & \multicolumn{1}{c|}{1155}            & \multicolumn{1}{c|}{1138}           & \textbf{769}   \\ \hline
\multicolumn{1}{|c|}{\multirow{4}{*}{BPIC2017}}     & all        & \multicolumn{1}{c|}{7616}           & \multicolumn{1}{c|}{\textbf{222}}    & \multicolumn{1}{c|}{6572}           & -              \\ \cline{2-6} 
\multicolumn{1}{|c|}{}                              & withCATE   & \multicolumn{1}{c|}{-}              & \multicolumn{1}{c|}{7595}            & \multicolumn{1}{c|}{7610}           & 5072           \\ \cline{2-6} 
\multicolumn{1}{|c|}{}                              & withoutCIW & \multicolumn{1}{c|}{-}              & \multicolumn{1}{c|}{3687}            & \multicolumn{1}{c|}{7594}           & \textbf{1707}  \\ \cline{2-6} 
\multicolumn{1}{|c|}{}                              & withoutTU  & \multicolumn{1}{c|}{\textbf{6535}}  & \multicolumn{1}{c|}{4933}            & \multicolumn{1}{c|}{\textbf{3723}}  & -              \\ \hline
\multicolumn{1}{|c|}{\multirow{4}{*}{TrafficFines}} & all        & \multicolumn{1}{c|}{12375}          & \multicolumn{1}{c|}{15193}           & \multicolumn{1}{c|}{12446}          & 15933          \\ \cline{2-6} 
\multicolumn{1}{|c|}{}                              & withCATE   & \multicolumn{1}{c|}{12513}          & \multicolumn{1}{c|}{12323}           & \multicolumn{1}{c|}{12277}          & 12325          \\ \cline{2-6} 
\multicolumn{1}{|c|}{}                              & withoutCIW & \multicolumn{1}{c|}{\textbf{12263}} & \multicolumn{1}{c|}{\textbf{12293}}  & \multicolumn{1}{c|}{\textbf{12262}} & \textbf{12312} \\ \cline{2-6} 
\multicolumn{1}{|c|}{}                              & withoutTU  & \multicolumn{1}{c|}{12530}          & \multicolumn{1}{c|}{12318}           & \multicolumn{1}{c|}{12325}          & 12335          \\ \hline
\end{tabular}
}
\label{tab:res1}
\end{table*}

\begin{table*}[hbtp]
\caption{Total gain (in Thousands) post-convergence for different proposal variants across various resource utilization levels and logs.}
\centering
\resizebox{0.75\linewidth}{!}{
\begin{tabular}{|cc|cccc|}
\hline
\textbf{}                                           & \textbf{}  & \multicolumn{4}{c|}{Resource utilization}                                                                                        \\ \hline
\multicolumn{1}{|c|}{Log}                           & Variant    & \multicolumn{1}{c|}{High}           & \multicolumn{1}{c|}{Moderately High} & \multicolumn{1}{c|}{Medium}        & Low            \\ \hline
\multicolumn{1}{|c|}{\multirow{4}{*}{BPIC2012}}     & all        & \multicolumn{1}{c|}{\textbf{6.8}}   & \multicolumn{1}{c|}{\textbf{1.6}}    & \multicolumn{1}{c|}{0.1}           & 0          \\ \cline{2-6} 
\multicolumn{1}{|c|}{}                              & withCATE   & \multicolumn{1}{c|}{3.8}            & \multicolumn{1}{c|}{0.5}             & \multicolumn{1}{c|}{0.3}           & 0.4            \\ \cline{2-6} 
\multicolumn{1}{|c|}{}                              & withoutCIW & \multicolumn{1}{c|}{4.3}            & \multicolumn{1}{c|}{0.1}             & \multicolumn{1}{c|}{0}             & 0              \\ \cline{2-6} 
\multicolumn{1}{|c|}{}                              & withoutTU  & \multicolumn{1}{c|}{5}              & \multicolumn{1}{c|}{0.1}             & \multicolumn{1}{c|}{\textbf{0.4}}  & \textbf{4.5}   \\ \hline
\multicolumn{1}{|c|}{\multirow{4}{*}{BPIC2017}}     & all        & \multicolumn{1}{c|}{1.1}            & \multicolumn{1}{c|}{\textbf{103.2}}  & \multicolumn{1}{c|}{14.3}          & 0              \\ \cline{2-6} 
\multicolumn{1}{|c|}{}                              & withCATE   & \multicolumn{1}{c|}{0}              & \multicolumn{1}{c|}{0.9}             & \multicolumn{1}{c|}{0.8}           & 45.8           \\ \cline{2-6} 
\multicolumn{1}{|c|}{}                              & withoutCIW & \multicolumn{1}{c|}{0}              & \multicolumn{1}{c|}{48.1}            & \multicolumn{1}{c|}{1.4}           & \textbf{78.5}  \\ \cline{2-6} 
\multicolumn{1}{|c|}{}                              & withoutTU  & \multicolumn{1}{c|}{\textbf{8.3}}   & \multicolumn{1}{c|}{47.8}            & \multicolumn{1}{c|}{\textbf{42.2}} & 0              \\ \hline
\multicolumn{1}{|c|}{\multirow{4}{*}{TrafficFines}} & all        & \multicolumn{1}{c|}{195.4}          & \multicolumn{1}{c|}{87.1}            & \multicolumn{1}{c|}{195.1}         & 81.2           \\ \cline{2-6} 
\multicolumn{1}{|c|}{}                              & withCATE   & \multicolumn{1}{c|}{197.6}          & \multicolumn{1}{c|}{194.8}           & \multicolumn{1}{c|}{\textbf{196}}  & \textbf{197.6} \\ \cline{2-6} 
\multicolumn{1}{|c|}{}                              & withoutCIW & \multicolumn{1}{c|}{\textbf{198.8}} & \multicolumn{1}{c|}{194.5}           & \multicolumn{1}{c|}{194.2}         & 194.8          \\ \cline{2-6} 
\multicolumn{1}{|c|}{}                              & withoutTU  & \multicolumn{1}{c|}{193}            & \multicolumn{1}{c|}{\textbf{196.8}}  & \multicolumn{1}{c|}{193}           & 190.2          \\ \hline
\end{tabular}
}
\label{tab:res2}
\end{table*}

Table~\ref{tab:res1} provides results related to the first component of RQ1, i.e., convergence rates of different variants of our proposal across diverse resource utilization levels and all logs. In the \textit{BPIC2012} and \textit{BPIC2017} logs, the \textit{all} variant demonstrates faster convergence under high or moderately high resource utilization conditions compared to the other variants. This means that the comprehensive information about significance, urgency, and capacity provided by the \textit{all} variant is particularly advantageous in scenarios where resources are relatively constrained. However, it is worth noting that when abundant resources are available, the \textit{all} variant does not converge as swiftly as other variants, especially when the number of cases the agent visits is relatively small. This discrepancy underscores the adaptability of our approach to varying resource availability levels.

Conversely, when examining the \textit{TrafficFines} log, it becomes apparent that all variants show a slightly significant difference in convergence rates. Interestingly, the \textit{withoutCIW} variant shows slightly faster convergence. This observation could be attributed to the characteristics of the TrafficFines log, which is notably larger in scale compared to the other two logs. It contains the highest number of cases and events. The large volume of cases and events in this log may lead to more predictable patterns, reducing the impact of specific information components, such as \textit{withoutCIW}, on the RL agent’s convergence. Therefore, the \textit{withoutCIW} variant might converge faster due to the log’s inherent characteristics.

Table~\ref{tab:res2} provides the results of the second part of~\ref{rq:rq1}, specifically, the total gain post-convergence. Similarly, when examining the \textit{BPIC2012}  log, it becomes evident that the \textit{all} variant outperforms other variants regarding total gain, particularly under high or moderately high resource utilization levels. The rationale behind these results is that when resources are relatively limited, the RL agent makes more informed and beneficial decisions, leading to superior total gains. In contrast, the \textit{withoutTU} variant performs comparatively better in scenarios with low resource utilization, suggesting that dismissing the total uncertainty might be more effective when available resources are high.

In the case of the \textit{BPIC2017} log, the \textit{all} variant achieves the highest total gain when resource utilization is at a moderately high level. However, it is noteworthy that under high resource utilization conditions, the RL agent initially achieves positive gains and converges relatively early. Nevertheless, it has been observed that the total gain drops below zero after 7,616 cases. Consequently, we report the total gain and convergence specifically after this 7,616 case threshold to provide a more comprehensive and accurate evaluation of the variant’s performance in this resource-intensive scenario. In the \textit{TrafficFines} log, \textit{all} variants show similar and indistinguishable performance. Still, the \textit{withCATE} variant outperforms others under medium and low resource utilization levels. This highlights the advantage of CATE in scenarios with large logs and medium to low resource utilization levels.


To address~\ref{rq:rq2}, we conduct a comparative analysis between the \textit{all} variant of our approach and two baseline methods, referred to as \textit{BL1}~\cite{bozorgicaise2023} and \textit{BL2}~\cite{metzger2020triggering}. Both BL1 and BL2 do not account for limited resources or uncertainty in their methods. We specifically selected the \textit{all} variant due to its superior performance across various logs and resource utilization levels, as it incorporates all the proposed dimensions introduced in this work. Furthermore, additional comparison results between various variants and the baseline methods can be found in the supplementary material\footnote{\url{https://doi.org/10.5281/zenodo.8352841}}.


  
  

\begin{figure}[hbtp]
  \includegraphics[width=\linewidth]{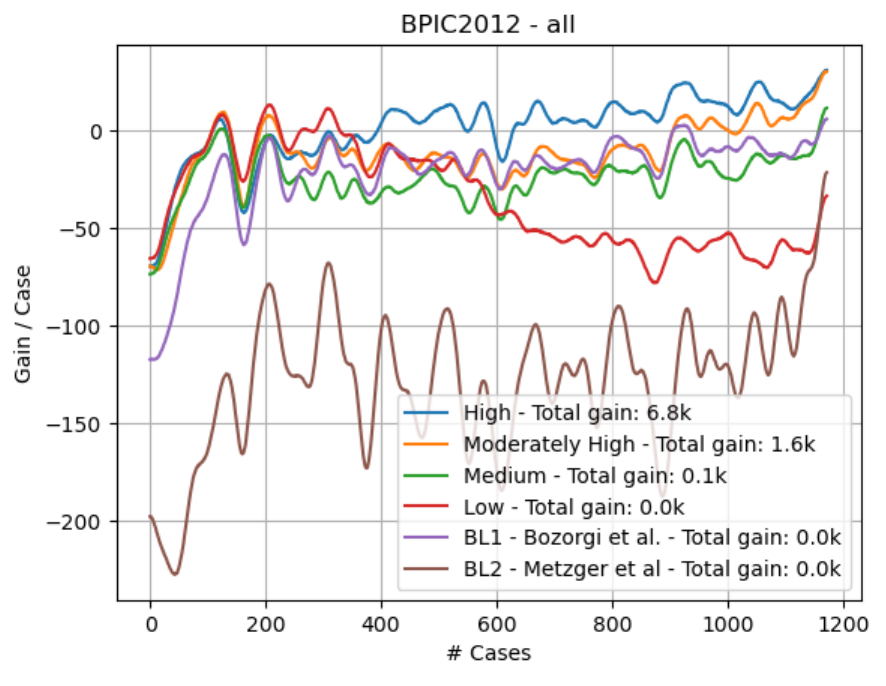}
  \caption{Comparative analysis of the all Variant and two baseline methods: \textit{BPIC2012}  }
  \label{fig:12}
\end{figure}

Regarding the \textit{BPIC2012} log, our approach demonstrates superior performance in terms of both convergence and total gain across diverse resource utilization levels compared to the baseline methods, as shown in Fig.~\ref{fig:12}. This observation underscores the effectiveness of our approach, particularly in resource-constrained scenarios, where it outperforms both baseline methods (BL1 and BL2). However, an exception arises at the low resource utilization level, characterized by abundant available resources. In this specific scenario, the RL agent takes considerably more time to converge and struggles to identify the optimal policy compared to other resource utilization levels. This suggests that, in resource-rich environments, one of the baseline methods, namely BL1, can exhibit competitive performance with our approach, while all other variants continue to outperform both baseline methods.

\begin{figure}[hbtp]
  \includegraphics[width=\linewidth]{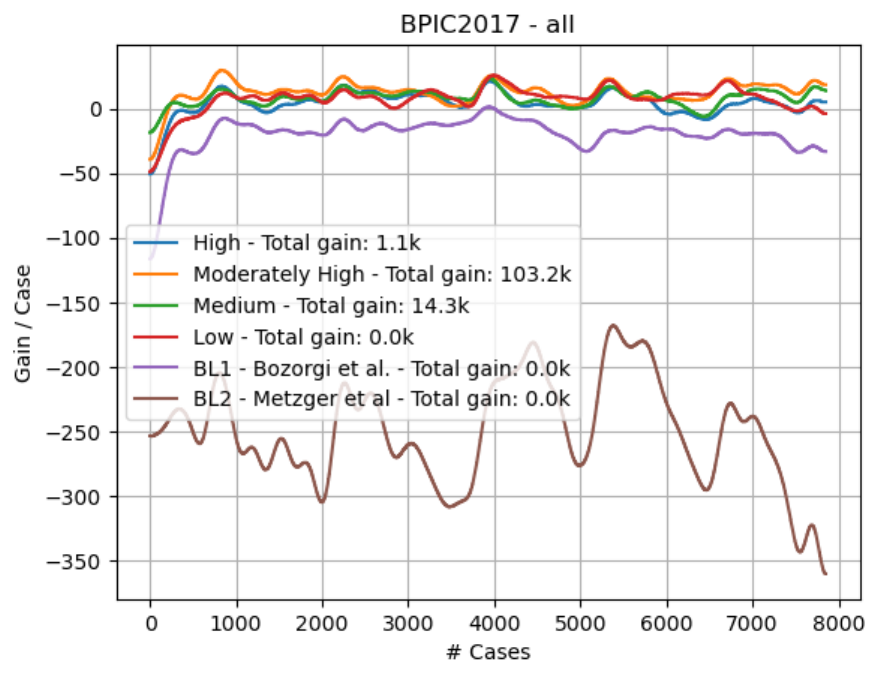}
  \caption{Comparative analysis of the all variants and two baselines: \textit{BPIC2017}. }
  \label{fig:17}
\end{figure}

On the contrary, the results from the \textit{BPIC2017} log demonstrate that our approach consistently outperforms both baseline methods (BL1 and BL2), as seen in Fig.~\ref{fig:17}. However, it is important to note that during the exploration phase of the RL agent’s learning process, we observed a scenario where, under high and medium resource utilization levels, the total gain dropped below zero after convergence. This means the RL agent could not reach a stable and positive-gain policy within the tested time frame.

\begin{figure}[hbtp]
  \includegraphics[width=\linewidth]{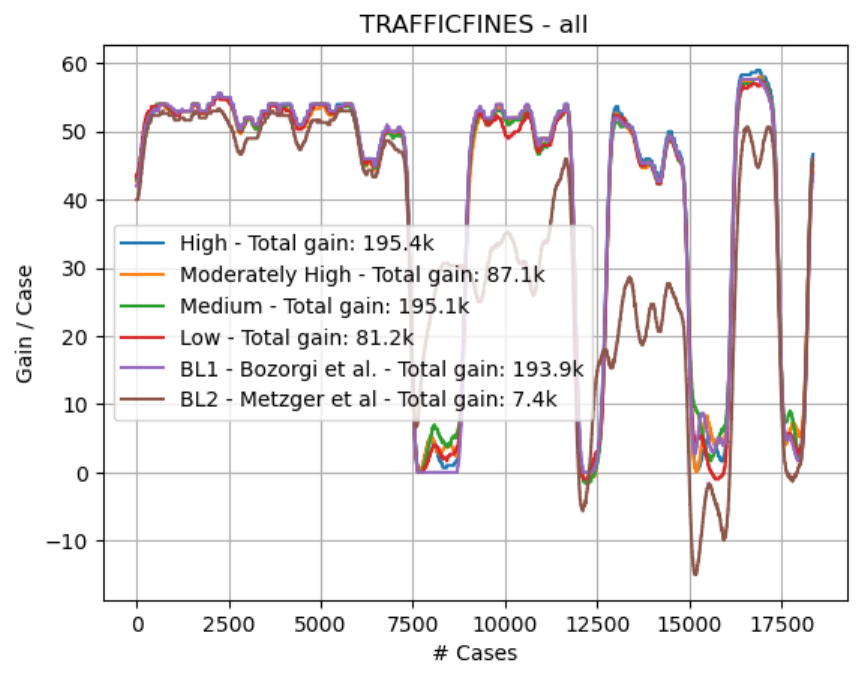}
  \caption{Comparative analysis of the all variants and two baselines: \textit{TrafficFines}. }
  \label{fig:traf}
\end{figure}

In the context of the \textit{TrafficFines} log, characterized by a substantial volume of cases and events, our approach and the baseline methods exhibit similar performance results, making them indistinguishable regarding their effectiveness, as seen in Fig.~\ref{fig:traf}. Several reasons contribute to this observed similarity. Firstly, the gain function considers the frequency of interventions triggered per case. However, the average case length in this event log is relatively short compared to the other two logs. Additionally, most cases result in positive outcomes, producing predominantly positive gains in the function’s returns. Secondly, as previously mentioned, the TrafficFines log contains a significantly more significant number of cases compared to the other two logs. This ample dataset provides the RL agent with sufficient cases to learn an intervention policy effectively, which may not be the case in real-world scenarios with limited data.

To summarize the findings from~\ref{rq:rq1}and~\ref{rq:rq2}, which are detailed above, 
it becomes evident that explicitly providing the RL agent with information regarding the significance, urgency, and capacity dimensions generally leads to a more effective intervention policy. In particular, the \textit{all} variant consistently performs well across different resource utilization levels compared to the baseline methods. This trend is particularly pronounced when resources are limited and there is a high level of uncertainty in the predictions. In such resource-constrained and uncertain environments, the advantages of supplying comprehensive information to guide the RL agent’s decision-making process become evident.

\subsection{Threats to Validity}

\paragraph{Internal Validity.} Potential threats to internal validity arise due to the stochastic nature of the RL agent’s learning process. To mitigate this threat, we conducted each experiment three times to address these uncertainties, and reported the middle point across the replications. Despite minor variations in convergence rates and total gains, we obtained consistent results across these replications. 


\paragraph{Ecological Validity.} Potential threats to the ecological validity of the findings arise from the assumption that all resources exhibit uniform proficiency in executing interventions. Additionally, the proposed approach assumes that there is only one type of intervention. In reality, there may be multiple types of interventions, such as contacting a customer to offer a discount, and offering a personalized consultation to the customer. The proposed method is not designed for such ``multi-intervention'' settings.


\paragraph{External Validity.} The generalizability of our evaluation is constrained by the utilization of only three datasets. The relatively low number of datasets is due to the fact that the experimental setup requires datasets where there is both a``case outcome'' and an ``intervention'', such that the intervention has a causal relation with the outcome. We reviewed all the datasets available in the 4TU Centre for Research Data\footnote{\url{https://data.4tu.nl/datasets/5ea5bb88-feaa-4e6f-a743-6460a755e05b}}) as well as public datasets used in previous studies on prescriptive process monitoring, but we were only able to identify three logs with the required characteristics. On the other hand, the three event logs come from different domains and have different characteristics.

Another threat to external validity comes from the fact that the study is based on event logs (sets of cases), wherein for each case in which an intervention is recorded, we only know the outcome given that the intervention occurred. We do not know what would be the outcome had the intervention not occurred. Vice versa, for cases where an intervention is not recorded for a case, we only know the outcome given that the intervention was not performed. We used a method for estimating the ``alternative outcome'' of each case (e.g. if an intervention occurred in a case, this method extrapolates the outcome should the intervention had not occurred). While the method we used for this purpose (RealCause) has a well-studied theoretical foundation and has been extensively evaluated~\cite{DBLP:journals/corr/abs-2011-15007}, the estimated alternative outcomes are not correct in all cases.




\section{Conclusion} \label{sec:conclusion}

This paper explored the hypothesis that integrating features related to significance (i.e.\ benefits) of an intervention, urgency of an intervention, and capacity to perform interventions, can enhance the total gain delivered by a reinforcement learning agent that triggers interventions to prevent negative case outcomes. The study examined this hypothesis within the constraints of limited resources available to perform interventions in the process and at different resource utilization levels.


The empirical evaluation highlights that various variants of the proposed approach, tailored for resource-constrained scenarios and accounting for uncertainties in the underlying business process, speed up the convergence towards an effective intervention policy. These variants consistently outperform existing baseline methods that do not consider resource limitations, both in terms of convergence and performance (total gain). The baseline methods tend to exhaust their resources prematurely, producing a suboptimal intervention policy. The enhancement in total gain achieved by the proposed method is higher when the resource capacity is more constrained. 




The evaluation also suggests that the use of conformal prediction techniques to model the uncertainty of the predicted factors given as input to the RL agent, helps the RL process to more quickly converge toward higher total gain.


Our proposal operates under the assumption that only a single type of intervention is available (for instance, offering a customer discount), and that this intervention is pre-defined based on domain knowledge. In real-world scenarios, a case may require multiple interventions of varying types (such as providing discounts, suggesting upgrades, offering vouchers for future purchases, etc.). An avenue for future research is to identify potential interventions based on empirical data, evaluate their effectiveness, and adapt our current methodology to accommodate a multi-intervention context. For instance, this could be achieved by leveraging multi-armed bandit approaches.

A promising avenue for future research involves tackling the challenges of a multi-objective RL task. In contrast to our current work, which focuses on enhancing a single objective, such as reducing the number of negative case outcomes (e.g., a customer rejecting a loan offer), the problem can be reframed to address multiple objectives simultaneously. For instance, it may involve reducing negative case outcomes while minimizing cycle time. To explore this, future research could investigate the use of multi-objective RL strategies, allowing for a more comprehensive and holistic approach to process optimization and decision-making in complex scenarios with competing objectives.



\paragraph{\textbf{Reproducibility}.} The supplementary material, which includes the necessary source code to replicate the experiments, is available at the following link: \url{https://github.com/mshoush/RL-prescriptive-monitoring} and \url{https://doi.org/10.5281/zenodo.8352841}


\paragraph{\textbf{Acknowledgments}.} 

This research is supported by the European Research Council (PIX Project). We acknowledge the use of \emph{ChatGPT}\footnote{OpenAI. (2023). ChatGPT (August 3 Version) [Large language model]. https://chat.openai.com} to enhance readability.




\bibliographystyle{splncs04} 
\bibliography{sample}

\end{document}